\pdfoutput=1

\documentclass[11pt]{article}

\usepackage[preprint]{acl}
\usepackage{times}
\usepackage{latexsym}

\usepackage[T1]{fontenc}

\usepackage[utf8]{inputenc}

\usepackage{microtype}

\usepackage{inconsolata}

\usepackage{graphicx}

\usepackage{hyperref}

\usepackage{amsmath}

\usepackage{amssymb}

\title{TDR: Task-Decoupled Retrieval with Fine-Grained LLM Feedback for In-Context Learning}

\author{
  Yifu Chen$^{3}$\thanks{~~Equal contribution.},
  Bingchen Huang$^{3}$\footnotemark[1],
  Zhiling Wang$^{3}$,
  Yuanchao Du$^{3}$,
  Junfeng Luo$^{3}$, \\
  \textbf{Lei Shen}$^{3}$\thanks{~~Corresponding author.},
  \textbf{Zhineng Chen}$^{1,2}$\footnotemark[2] \\
  $^1$School of Computer Science, Fudan University \\
  $^2$Shanghai Collaborative Innovation Center of Intelligent Visual Computing \\
  $^3$Meituan \\
  \texttt{shenlei1996@gmail.com}, 
  ~\texttt{zhinchen@fudan.edu.cn} \\
  \texttt{\{chenyifu05, huangbingchen, wangzhiling02, duyuanchao, luojunfeng\}@meituan.com} \\
}

\begin{document}
\maketitle
\begin{abstract}
In-context learning (ICL) has become a classic approach for enabling LLMs to handle various tasks based on a few input-output examples. The effectiveness of ICL heavily relies on the quality of these examples, and previous works which focused on enhancing example retrieval capabilities have achieved impressive performances. However, two challenges remain in retrieving high-quality examples: (1) Difficulty in distinguishing cross-task data distributions, (2) Difficulty in 
making the fine-grained connection between retriever output and feedback from LLMs. In this paper, we propose a novel framework called TDR. TDR decouples the ICL examples from different tasks, which enables the retrieval module to retrieve examples specific to the target task within a multi-task dataset. Furthermore, TDR models fine-grained feedback from LLMs to supervise and guide the training of the retrieval module, which helps to retrieve high-quality examples. We conducted extensive experiments on a suite of 30 NLP tasks, the results demonstrate that TDR consistently improved results across all datasets and achieves state-of-the-art performance. Meanwhile, our approach is a plug-and-play method, which can be easily combined with various LLMs to improve example retrieval abilities for ICL. The code is available at \url{https://github.com/Nnn-s/TDR}.
\end{abstract}

\section{Introduction}

Large language models (LLMs) like GPT-4\cite{openai2024gpt4technicalreport} have demonstrated exceptional performance across a wide range of language tasks. These models are typically trained on vast datasets, implicitly storing a significant amount of world or domain knowledge within their parameters. However, they are also prone to hallucinations and cannot fully represent long-tail knowledge from their training corpora\cite{xie2021explanation}. In-context learning (ICL)\cite{brown2020language, black2021gpt, luo2023dr} has emerged as a transformative approach for LLMs, enabling them to effectively leverage long-tail knowledge learned during training with minimal input-output examples, thereby significantly reducing model hallucinations without requiring any updates to model parameters. The effectiveness of ICL heavily depends on the quality of the provided examples\cite{liu2021makes}. 
As proposed by \cite{wang2023learning} and \cite{shi2023replug}, the task of retrieving in-context examples for LLMs is specifically designed to improve the quality of retrieved examples. Our work builds on these foundations and focuses on enhancing the retrieval capability of high-quality in-context examples to maximize the potential and performances of LLMs.


\begin{figure}[t]
  \includegraphics[width=\columnwidth]{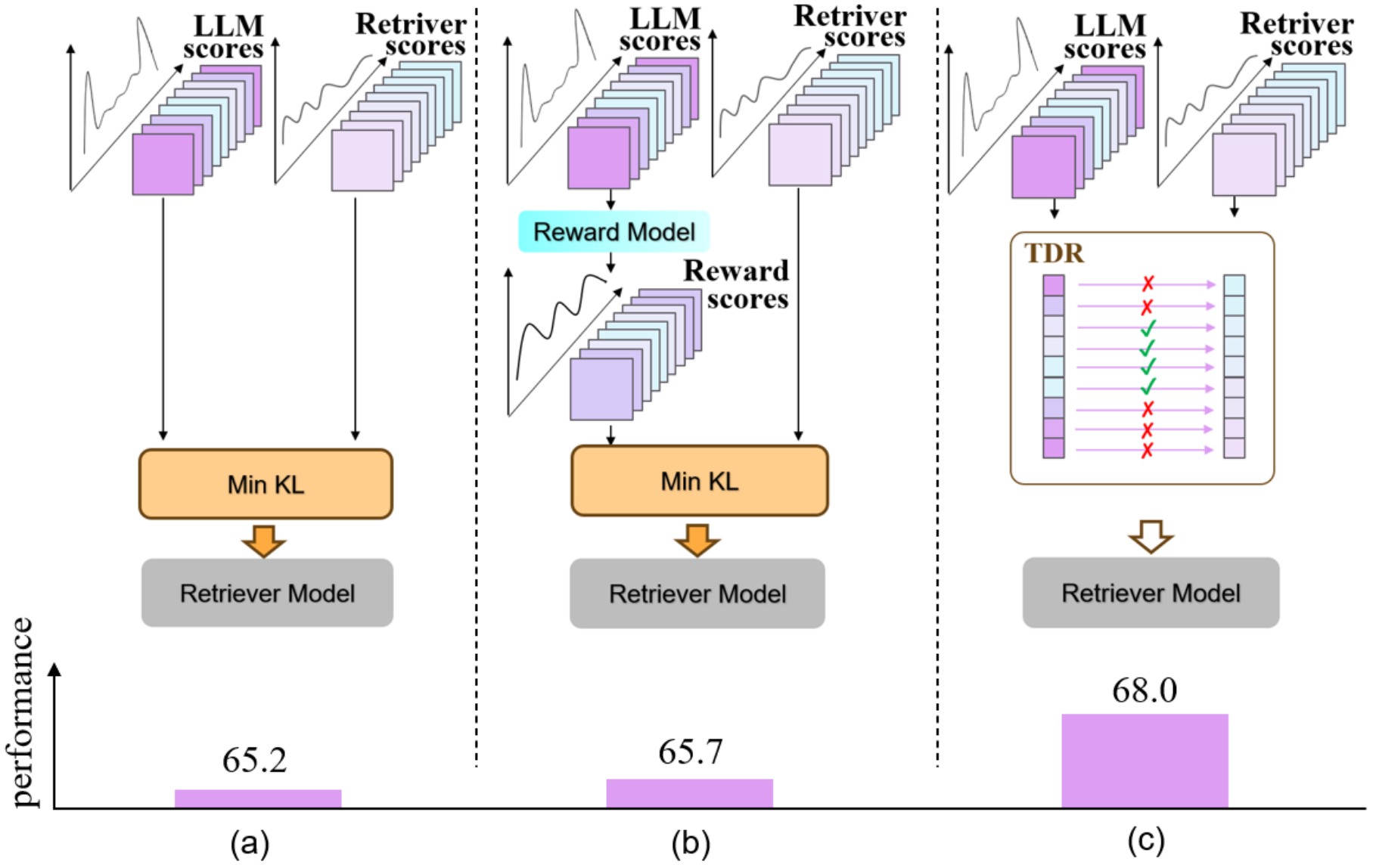}
  \caption{Comparison with previous methods. (a) KL divergence-based method: Uses LLM scores with KL divergence minimization, Performance is limited by the large distributional gap between retriever scores and LLM scores (b) Reward model-based KL method: Applies a reward model to smooth scores but still uses KL divergence, improving performance over (a) while facing similar alignment challenges. (c) Our method: Selects retrieval candidates using LLM scores, establishing positive correlation without distribution fitting, thus avoiding misalignment and improving performance.}
  \label{fig:comparation}
\end{figure}

Despite these advances, several challenges remain to understand and improve the effectiveness of ICL, which limits its potential. One such challenge is distinguishing data from different tasks. In real-world scenarios, retrieval pools often contain examples from multiple tasks, with significant differences in data distribution and characteristics. Retrieving examples from other tasks can negatively impact LLMs learning from in-context examples. However, this challenge is barely investigated in previous work. 
Table \ref{table:diff_task_case_analyse} in the Appendix shows specific examples retrieved from other tasks, which have texts similar to the query and significantly different answer patterns, making it difficult for LLMs to learn from these retrieval examples.

Another challenge is how to make the fine-grained connection between retriever output and feedback from LLMs. The relationship between the scores output by retriever and LLM feedback scores can be highly correlated. The retriever trained with LLM feedback exhibits a more consistent scoring pattern when compared to the LLM feedback scores\cite{wang2023learning}. In contrast, the scatter distribution of E5\cite{wang2022text} which is not trained with fine-grained LLM feedback shows greater fluctuation and instability. It is crucial to establish a direct and efficient relationship between the output of retriever and LLM to enhance the quality of retrieved samples.


In this paper, we propose a novel framework for retrieving high-quality in-context examples for large language models, named TDR. We start with a bi-encoder\cite{devlin2018bert} as the initial dense retriever to obtain a candidate set of examples. 
By decoupling the training of examples from different tasks, TDR enable the retriever to focus on retrieving relevant data specific to the target task within a multi-task dataset, thereby improving the precision and relevance of retrieved examples. 
Besides, TDR employs a specific loss function TDR to model the fine-grained feedback from LLMs and guide the training of the dense retriever. This process can be iterated multiple times to enhance the retriever's ability to retrieve high-quality examples from the specific task.

Following the task setting of \cite{wang2023learning}, we conducted experiments on a dataset comprising 30 diverse NLP tasks, spanning nine categories including question answering, natural language inference, commonsense reasoning, and summarization, etc. Extensive experimental results obtained using LLaMA-7B \cite{touvron2023llama} demonstrate that our method outperforms the previous state-of-the-art approach, showing consistent improvements in in-context learning performance across all tasks. Similar gains are observed for unseen tasks during training and across LLMs of varying sizes, further validating the effectiveness and versatility of our strategy.

Contributions of this paper can be summarized as follows: 

-We analyze the key factors affecting the capabilities of retrieving in-context examples for large language models and observe that distinguishing data from different tasks and making fine-grained connection between the outputs of retriever and LLMs count most. 

-We propose TDR, a novel scheme to promote retrieving high-quality contextual examples for large language models. Specifically, decoupling the training of examples from different tasks is developed to further distinguishing data from different domains. Meanwhile, we employ a correlation-enhanced loss function to model the fine-grained feedback from LLMs, which can make better use of feedback from LLMs. 

-Extensive evaluation on 30 NLP tasks demonstrates that TDR outperforms previous state-of-the-art method, achieving a state-of-art performance across all tasks including seen and unseen tasks during training. 

\begin{figure*}[t]
  \includegraphics[width=0.95\linewidth]{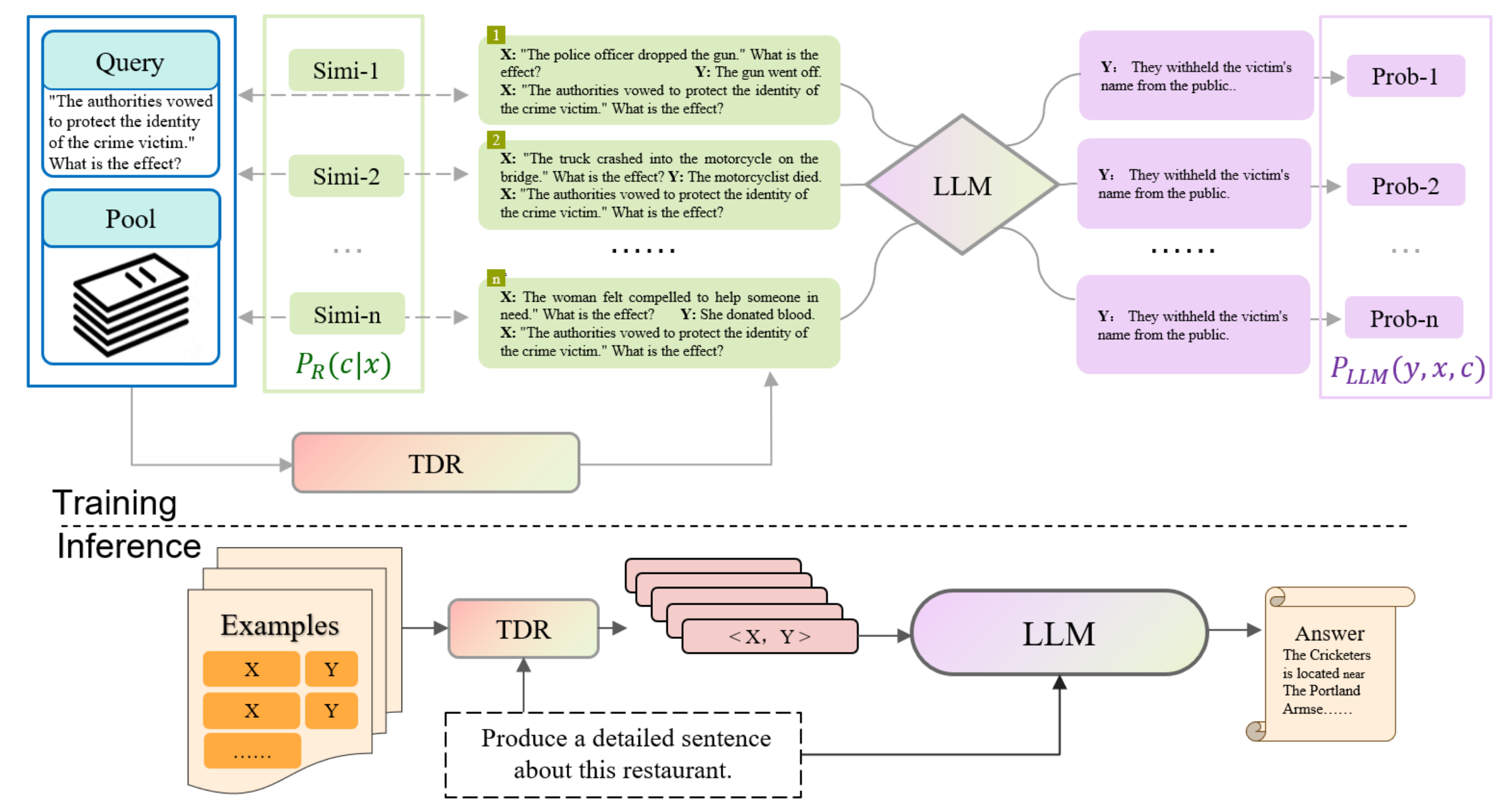}
  \caption{TDR Framework for Retriever Fine-Tuning and Inference. Training: The retriever selects task-specific examples based on queries, while the LLM generates corresponding probabilities. TDR optimizes the retriever to maximize the likelihood of correct answers given queries and examples (Section~\ref{subsec:Lce}). Inference: The fine-tuned retriever retrieves in-context examples from pool $\mathbb{P}$, which are concatenated with the query and fed to the LLM for prediction.}
  \label{fig:method_framework}
\end{figure*}

\section{Related Work}

\subsection{In-context learning}

In-context learning (ICL) is an emergent capability of large language models (LLMs) that allows them to solve tasks by conditioning on input-output demonstrations without parameter updates. This phenomenon has been widely studied in models like GPT-3\cite{brown2020language}, PaLM\cite{chowdhery2023palm}, and LLaMA\cite{touvron2023llama}. Research on ICL primarily focuses on two directions: mechanistic interpretation and example optimization strategies.

For mechanistic understanding, Studies\cite{xie2021explanation} proposes diverse theoretical frameworks and interprets ICL as implicit Bayesian inference, where models update latent task representations based on demonstrations. Concurrently, \cite{von2023transformers} argues that transformers implicitly perform gradient descent during ICL, mimicking meta-optimization processes. Recent work\cite{park2024iclr} further reveals that LLMs dynamically reconfigure semantic representations when contextual examples scale, shifting from pretrained priors to task-specific structures.

In example optimization, researchers explore strategies to enhance ICL performance through prompt engineering and data selection. Retrieval-based methods, such as BM25-based selection\cite{reimers2019sentence} and contrastive retrievers\cite{rubin2021learning}, aim to identify semantically relevant examples. Advanced techniques like determinantal point processes\cite{ye2023compositional} model inter-example interactions, while structured prompting\cite{hao2022structured} extends context length to thousands of tokens. The LLM-R framework\cite{wang2023learning} introduced a novel approach using a reward model to iteratively train dense retrievers for identifying high-quality in-context examples. Our work aligns with this direction, proposing a novel method for dynamic example selection.

\subsection{Retrieval-augmented Models}
Retrieval-augmented large language models (RALMs) integrate generative capabilities with external knowledge to enhance factual accuracy and timeliness\cite{guu2020retrieval, borgeaud2022improving}. This paradigm addresses hallucinations and outdated knowledge in LLMs while enabling source attribution\cite{lewis2020retrieval}. Methods like \cite{guu2020retrieval, borgeaud2022improving} pretrain retrievers jointly with LLMs, encoding retrieved documents into latent representations for generation. Alternatively, kNN-LM\cite{khandelwal2019generalization} interpolate model predictions with retrieved token distributions. While kNN-LM avoids additional training, it still requires access to internal model representations. Recently, the utilization of feedback from LLMs received attention from researchers, \cite{shi2023replug} directly applies LLM probabilities as LLM feedback. While \cite{wang2023learning} introduced a novel approach to iteratively train dense retrievers for identifying high-quality in-context examples, studies have shown that training retrievers to leverage fine-grained LLM feedback significantly enhances in-context learning performance compared to traditional methods like BM25\cite{reimers2019sentence} that do not utilize such feedback.

\section{Proposed Method}
In this section, we introduce the training pipeline of our method as illustrated in Figure \ref{fig:method_framework}, including architecture, training data generation, correlation-enhanced loss, task-mask mechanism.

\subsection{Architecture}
\paragraph{Retriever}
We adopt a bi-encoder based dense retriever architecture initialized with $\text{E5}_\text{{base}}$  due to its excellent performance. Given a query $x$ and the candidate examples $\{c_i\}_{i = 1}^{n}$, our retriever encodes the query $x$ into an embedding $E(x)$ and each of the candidate examples into embeddings $E(c_i)$. The retriever score between the query and each example is computed via the dot product:

\begin{equation}
    \label{eq:cos_sim}
    s(x,c_i)=E(x)\cdot E(c_i),
\end{equation}


\paragraph{Large Language Model}
To make a fair comparison with other existing approaches, we opt specifically for LLAMA \cite{touvron2023llama}.

\subsection{Training data generation}


For each training example $(x, y)$, we retrieve top-$n$ candidates $\{(x_i, y_i)\}_{i=1}^{n}$ from a diverse pool $P$, excluding $(x, y)$. Candidates are represented as $(x_i, y_i)$, and retrieval is based on $x$. The candidates are ranked using a frozen LLM by computing the logarithm of the conditional probability of $y$ given $x$ and each candidate $(x_i, y_i)$:
\begin{equation}
    \label{eq:llm_score}
    \begin{aligned}
        P_{LLM}(y , c_i,x) &=  Task(\  \log p_{llm}(y | x, c_i)\ ),\\
        \log p_{llm}(y|x, c_i) &= \sum_{j=1}^{n} \log p_{llm}(y_j|x, c_i, y_{<j}),
    \end{aligned}
\end{equation}
where $Task()$ assigns a low score if $c_i$ is from a different task than $x$. This method requires only a single forward pass, making it computationally efficient and task-agnostic.

\begin{table*}[t]
\centering
\begin{tabular}{lllllllllll}
\hline
\# of datasets→ & CQA           & Comm.         & Coref.        & NLI           & Para.         & RC            & Sent.         & D2T           & Summ.         & Avg           \\
task number                          & 3             & 3             & 3             & 5             & 3             & 4             & 3             & 3             & 3             & 30            \\ \hline
Zero-shot                                           & 29.0          & 71.5          & 66.8          & 44.0          & 60.0          & 41.3          & 50.5          & 25.6          & 17.5          & 44.9          \\
Random                                              & 40.4          & 77.6          & 67.2          & 50.9          & 56.6          & 58.1          & 88.8          & 47.0          & 38.9          & 57.9          \\
K-means                                             & 41.6          & 79.5          & 66.0          & 50.8          & 52.6          & 53.6          & 90.9          & 42.5          & 40.5          & 57.0          \\
BM25                                                & 45.9          & 78.1          & 62.9          & 54.7          & 66.1          & 59.9          & 89.6          & 49.3          & 50.0          & 61.3          \\
E5$_{base}$                                              & 49.0          & 79.8          & 64.6          & 53.6          & 58.0          & 60.2          & \textbf{94.4} & 48.0          & 50.0          & 61.4          \\
SBERT                                               & 48.5          & 79.3          & 64.2          & 57.5          & 64.1          & 60.6          & 91.9          & 47.4          & 49.3          & 62.1          \\
EPR                                                 & 48.4          & 79.3          & 64.4          & 64.3          & 65.1          & 59.8          & 91.7          & 49.7          & 50.0          & 63.5          \\
LLM-R                                               & 48.7          & 80.4          & 70.4          & \textbf{72.5} & 71.5          & 59.0          & 93.6          & \textbf{49.9} & 51.1          & 66.5          \\ \hline
Ours(1 iter)                                        & \textbf{55.2} & 80.1          & 64.7          & 71.3          & 80.8          & \textbf{65.0} & 92.2          & \textbf{49.9} & \textbf{51.3} & 68.0          \\
Ours(2 iter)                                        & 55.1          & \textbf{80.5} & 69.1          & 71.0          & 81.9          & 64.3          & 92.1          & 49.3          & \textbf{51.3} & \textbf{68.3} \\
Ours(3 iter)                                        & 54.5          & 79.9          & \textbf{70.5} & 71.5          & \textbf{82.2} & 63.5          & 90.4          & 49.0          & 51.1          & 68.1          \\ \hline
\end{tabular}
\caption{\label{main-results}
 Main results on a suite of 30 NLP tasks. Other results come from \cite{wang2023learning}.
}
\end{table*}

\subsection{Correlation-enhanced Loss}
\label{subsec:Lce}
To provide fine-grained supervision for the retriever based on LLM probabilities, we propose a novel \textbf{correlation-enhanced loss}. This loss function is designed to align the retriever's behavior with the language model's preferences by explicitly modeling the relationship between retrieval likelihoods and LLM probabilities. In the following, we detail the computation of our proposed loss function.

\subsubsection{Probabilities of the retrieved examples}
Each candidate example $c_i$ is selected according to its similarity score $s(x, c_i)$ with respect to the query $x$, where $\{s(x, c_i)\}_{i=1}^{n}$ represents the set of similarity scores for the top-$n$ candidates. These scores serve as the foundation for computing the retrieval likelihood. Specifically, the retrieval likelihood for each candidate $c_i$ is calculated as:
\begin{equation}
    \label{eq:P_R}
    P_{R}(c_i \mid x)  = \frac{e^{s(x, c_i) / \gamma}}{\sum_{c_j \in \mathcal{D}^{\prime}} e^{s(x, c_j) / \gamma}},
\end{equation}

where $\gamma$ is a hyperparameter that controls the temperature of the softmax. This retrieval likelihood reflects the retriever's confidence in the relevance of each candidate example to the query. Ideally, the retrieval likelihood should be computed by marginalizing over all examples in the corpus $\mathcal{D}$, but this is computationally intractable in practice. Therefore, we approximate the retrieval likelihood by marginalizing only over the retrieved candidate examples $\mathcal{D}^{\prime}$. And also in our framework, since the retrieval results are pre-computed, we avoid the need to encode the entire corpus during training.

\subsubsection{Align probabilities}
To align the retriever's behavior with the language model's preferences, we utilize pre-computed LLM probabilities derived from the previously constructed dataset. For each candidate example $c_i \in \mathcal{D}^{\prime}$, where $\mathcal{D}^{\prime}$ denotes the set of retrieved candidates, we employ the pre-computed probability $P_{LLM}(y, c_i, x)$ as defined in Equation \ref{eq:llm_score}. This probability quantifies the likelihood of the ground truth output $y$ given the input context $x \in \mathcal{B}$ and the candidate example $c_i$. These probabilities are computed using a frozen language model during the dataset construction phase, ensuring consistency and efficiency in training.

The correlation-enhanced loss of a batch of candidates $c$ is defined as the element-wise product of two components: (1) the retrieval likelihood $P_{R}(c \mid x)$, and (2) the pre-computed LLM probability $P_{LLM}$. Formally, the loss is expressed as:  

\begin{equation}
    \label{eq:Q_CE}
    Q_{CE}(c, x, y) = \sum_{c_i \in \mathcal{D}^{\prime}}P_{R}(c_i \mid x) \cdot P_{LLM}(y, c_i, x),
\end{equation}

This formulation ensures that examples with high LLM probabilities are prioritized during training. The training objective is to optimize the retriever to prioritize candidates with the highest $P_{LLM}(y \mid c, x)$ for better LLM predictions, which is achieved by minimizing the following loss function:

\begin{equation}
    \label{eq:L_CE}
    \mathcal{L}_{CE} = -\frac{1}{|\mathcal{B}|} \sum_{x \in \mathcal{B}}   Q_{CE}(c, x, y),
\end{equation}

where $\mathcal{B}$ is a batch of input contexts. By minimizing this loss, we encourage the retriever to prioritize examples that are not only relevant to the input context but also beneficial for the language model's predictions.

\subsubsection{Explanation of $L_{CE}$ Mechanism}
For the low-LLM-probability candidate $c_i$, the term $P_{LLM}(y, c_i, x)$ results in a very large negative probability score because $p_{llm}(y \mid x, c_i)$ is a positive number close to zero. Furthermore, for candidate $c_i$ originating from different tasks, the function $Task(\cdot)$ in Equation \ref{eq:llm_score} assigns a similarly large negative probability score. Both of these situation compel the retriever to decrease the cross-entropy loss $L_{CE}$ by reducing $P_R(c_i \mid x)$, achieved by lowering $s(x, c_i)$. Due to the Softmax redistribution mechanism in Equation \ref{eq:P_R}, adjusting $s(x, c_i)$ for one candidate automatically redistributes probability mass across all candidates $\{c_1,...,c_m\}$, ensuring that $\sum_{n=1}^{m} P_R(c_n \mid x) = 1$. Consequently, for candidates $c_j$ that have high LLM-probability scores, $P_R(c_j \mid x)$ (and correspondingly $s(x , c_j)$) will increase to offset the decrease in $P_R(c_i \mid x)$.

\subsection{Task-Mask Mechanism}
$\mathcal{L}_{CE}$ solves the problem of aligning probabilities between our retriever and the LLM, but a crucial issue is observed. Specifically, when calculating $\mathcal{L}_{CE}$ , examples from different tasks are inherently assigned very large negative values which results in disproportionately high loss values compared to those from the same task. It aids our retriever in learning to penalize the selection of examples from different tasks, but hinders its ability to find more suitable examples within the same task.

To mitigate this issue, we design a Task-Mask Mechanism that separates the loss computation by introducing loss mask $\mathcal {M} \in \mathbb{R}^\mathcal{B}$:

\begin{equation}
    \begin{aligned}
        \mathcal{M} &= \{ \mathcal{M}_1, \mathcal{M}_2, \cdots, \mathcal{M}_\mathcal{B} \},\\
        \mathcal{M}_x &= \begin{cases} 
            1, & \text{if  } \  p_{min} < t \\
            0, & \text{otherwise}
            \end{cases},\quad x \in \mathcal{B} \\
    \end{aligned}
\end{equation}

Here, $t$ denotes the task threshold, a large negative value, and $\{\}$ signifies the concatenation operation. The term $ p_{min} \in \mathbb{R}^1$ denotes the minimum of $P_{LLM}$ with a single batch. $\mathcal{L}_{CE}$ is then divided into two components: the different-task loss $\mathcal{L}_d$, which discourages retrieving from different tasks, and the same-task loss $\mathcal{L}_s$, which encourages retrieving better examples within the same task:

\begin{equation}
    \begin{aligned}
        \mathcal{L}_{d} &= -\frac{1}{|\mathcal{B}|} \sum_{x \in \mathcal{B}} Q_{CE}(c, x, y) \cdot \mathcal{M}_x, \\
        \mathcal{L}_{s} &= -\frac{1}{|\mathcal{B}|} \sum_{x \in \mathcal{B}} Q_{CE}(c, x, y) \cdot (1 - \mathcal{M}_x),
    \end{aligned}
\end{equation}

And then in alignment with \cite{wang2023learning}, we integrate an InfoNCE-based contrastive loss $\mathcal{L}_{cont}$ \cite{chen2020simple} to incorporate the in-batch negatives by designing the candidate with the highest LLM probabilities as the positive example. Thus, the final training objective for the retriever can be formally expressed as:
\begin{equation}
     \mathcal{L}_{retriever}= \lambda \cdot\mathcal{L}_{cont} + \alpha \cdot\mathcal{L}_{d} + \beta \cdot \mathcal{L}_{s},
\end{equation}
where \{$\lambda$, $\alpha$, $\beta$\} are the hyperparameters that determine the relative weighting of the three loss functions.

\section{Experiments}

\subsection{Evaluation Setup}
Following the task setting of \cite{wang2023learning}, we verify the merit of the proposed TDR for a diverse collection of 30 publicly available NLP tasks\cite{wei2021finetuned,cheng2023uprise,wang2023learning}, which span 9 distinct categories and include up to 10k examples per dataset.
The training retrieval pool is constructed by combining all training examples, excluding the four datasets QNLI, PIQA, WSC273, and Yelp, aiming to assess the models' generalization ability on unseen tasks. Detailed task classification is shown in Table \ref{table:datasets}.

\begin{table*}
\centering
\begin{tabular}{llllll}
\hline
Category      & \multicolumn{5}{l}{Datasets}                                  \\ \hline
Close QA      & ARC Challenge & ARC Easy  & NQ              &          &      \\
Commonsense   & COPA          & HellaSwag & \textbf{PIQA}   &          &      \\
Coreference   & Winogrande    & WSC       & \textbf{WSC273} &          &      \\
Paraphrase    & MRPC          & PAWS      & QQP             &          &      \\
Sentiment     & Sentiment140  & SST2      & \textbf{Yelp}   &          &      \\
Data-to-text  & CommonGen     & DART      & E2E NLG         &          &      \\
Summarize     & AESLC         & AGNews    & Gigaword        &          &      \\
Reading Comp. & BoolQ         & MultiRC   & OpenBook QA     & SQuAD v1 &      \\
NLI           & MNLI (m)      & MNLI (mm) & \textbf{QNLI}   & RTE      & SNLI \\ \hline
\end{tabular}
\caption{\label{table:datasets}
 Detailed datasets used in this paper. The bold texts display four held-out datasets which are unseen during training periods.
}
\end{table*}

\begin{table*}
\centering
\begin{tabular}{ccccccccccccc}
\hline
    \#      &$\mathcal{L}_{CE}$& Task-Mask  & CQA  & Comm. & Coref. & NLI  & Para. & RC   & Sent. & D2T  & Summ. & Avg \\
\hline
    1       &                  &      &     48.0  &   79.4   &  \textbf{67.0}   &   67.0   &   74.0    &  60.5    &  91.5     & 49.6     &  50.3     & 65.2 \\
    2       &$\checkmark$      &      &    54.7   &   79.6     & 66.0 &   71.2    &   76.4   &   63.4    &  91.4    &   \textbf{50.2}    & \textbf{51.3} & 67.3\\
    3       &$\checkmark$      &$\checkmark$     & \textbf{55.2} &\textbf{ 80.1 }         & 64.7          & \textbf{71.3 }         & \textbf{80.8}          & \textbf{65.0} & \textbf{92.2 }         & 49.9 & \textbf{51.3} & \textbf{68.0}    \\
\hline
\end{tabular}
\caption{\label{ablation}
Ablation study of our proposed TDR on the test set. The values in the table show the average performance of the model across 9 categories consisting of 30 tasks.
}
\end{table*}


During training, we initialize the retriever using the pre-trained $\text{E5}_\text{{base}}$ model \cite{wang2022text}. The retriever is fine-tuned on the generated dataset with a batch size of 32 and 4 examples per batch. Training is conducted for 12,000 steps on 8 V100 GPUs, completing in approximately two hours, with a learning rate of $3 \times 10^{-5}$. To mitigate the influence of random seeds, we report the average performance metrics across each task category. For task evaluation, we employ LLaMA-7B \cite{touvron2023llama} as the standard language model to ensure consistency and fairness in comparisons. Following prior work \cite{wang2023learning}, we retrieve 8 in-context examples for each test input in all evaluations except zero-shot settings.

Building upon this foundation, our method TDR addresses the insufficient utilization of LLM feedback in complex training procedures by explicitly modeling LLM-generated feedback to supervise retriever training. Additionally, we decouple the training of examples across distinct tasks, further enhancing performance across all evaluated tasks. We perform three iterative training cycles, as the second iteration yields the best performance. The experimental results are recorded as "Ours 1 iter," "Ours 2 iter," and "Ours 3 iter" in Table \ref{main-results}. The results demonstrate that our approach achieves significant improvements across seven task categories, delivering an average accuracy gain of 1.8\% over the previous state-of-the-art method. Notably, TDR surpasses previous SOTA method by 10.7\% on the task category Paraphrase, validating its significant effectiveness.

Furthermore, as shown in Figure \ref{fig:performance_gains}, our method significantly outperforms the "Random" baseline, achieving an average improvement of 22.2\% across all 30 tasks, highlighting its effectiveness in leveraging task - specific information. It also demonstrates robust generalization, consistently beating the random baseline on four unseen training tasks, indicating its ability to handle open - set scenarios. However, it performs relatively poorly on the WSC and RTE tasks, likely due to the limited number of training examples (554 for WSC and 2,490 for RTE) in a 600,000 - example retrieval pool, which may impede the retriever. Despite this, our method still yields competitive results, showing its robustness across diverse tasks.

Detailed experimental results for all 30 tasks are provided in Table \ref{table:detailed_results} of the supplementary material. In the subsequent experiments, we consistently refer to our method as TDR, which corresponds to the "Ours 2 iter" configuration.

\section{Analysis}

\subsection{Ablation Study}
Here, we study how each component in TDR influences the overall performance. We consider one or more components at each stage and Table \ref{ablation} summarizes the results on training set of the 9 categories consisting of 30 NLP tasks. Note that baseline at Row \#1 is a dense bi-encoder retriever finetuned by minimize the KL-Divergence between the retriever score distribution and the LLM preference. 

By incorporating the $\mathcal{L}_{CE}$ that appropriately aligns the retriever probabilities $P_R$ and task-specific LLM probabilities $P_{LLM}$, the variant at Row \#2 makes the absolute improvement over the base model at Row \#1 on the average score. This is not surprised as the correlation-enhanced loss $\mathcal{L}_{CE}$ can establish a positive correlation between the retriever probabilities and the LLM probabilities while avoiding the direct use of KL divergence to fit the distributions, given the significant differences between them. Specifically, the retriever actively adjusts its vector space to bring the example $c$, which maximizes the probability of the answer $y$, closer to the given query $x$. 

The task-mask mechanism further enhances the retriever by dividing the training objective into two parts: distinguishing between different tasks and finding better examples within the same task, achieving the best results as shown in Row \#3.

\begin{table*}[t]
\centering
\begin{tabular}{lllllllllll}
\hline

             & CQA  & Comm. & Coref. & NLI  & Para. & RC   & Sent. & D2T  & Summ. & Avg           \\ \hline
gpt-neo-2.7b &      &       &        &      &       &      &       &      &       &               \\
BM25         & 41.1 & 67.0  & 53.2   & 47.6 & 64.5  & 51.2 & 78.3  & 45.4 & 47.3  & 54.4          \\
LLM-R        & 42.2 & 68.0  & 59.7   & 71.5 & 73.0  & 51.6 & 91.6  & 46.9 & 48.8  & 61.8          \\
Ours         & 41.4 & 67.8  & 60.4   & 70.2 & 82.0  & 53.4 & 90.9  & 46.0 & 48.8  & \textbf{62.3} \\ \hline
llama-13b    &      &       &        &      &       &      &       &      &       &               \\
BM25         & 49.6 & 80.1  & 61.1   & 67.0 & 69.9  & 60.5 & 92.5  & 49.9 & 50.9  & 64.6          \\
LLM-R        & 52.0 & 83.7  & 71.2   & 76.8 & 73.3  & 62.2 & 94.2  & 50.7 & 52.0  & 68.8          \\
Ours         & 59.2 & 83.3  & 70.4   & 74.3 & 82.2  & 64.6 & 93.2  & 49.8 & 51.9  & \textbf{69.9} \\ \hline
\end{tabular}
\caption{\label{various-llms-results}
Generalization to LLMs that are not used for training.
}
\end{table*}

\begin{figure*}[t]
    \centering
  \includegraphics[width=0.85\linewidth]{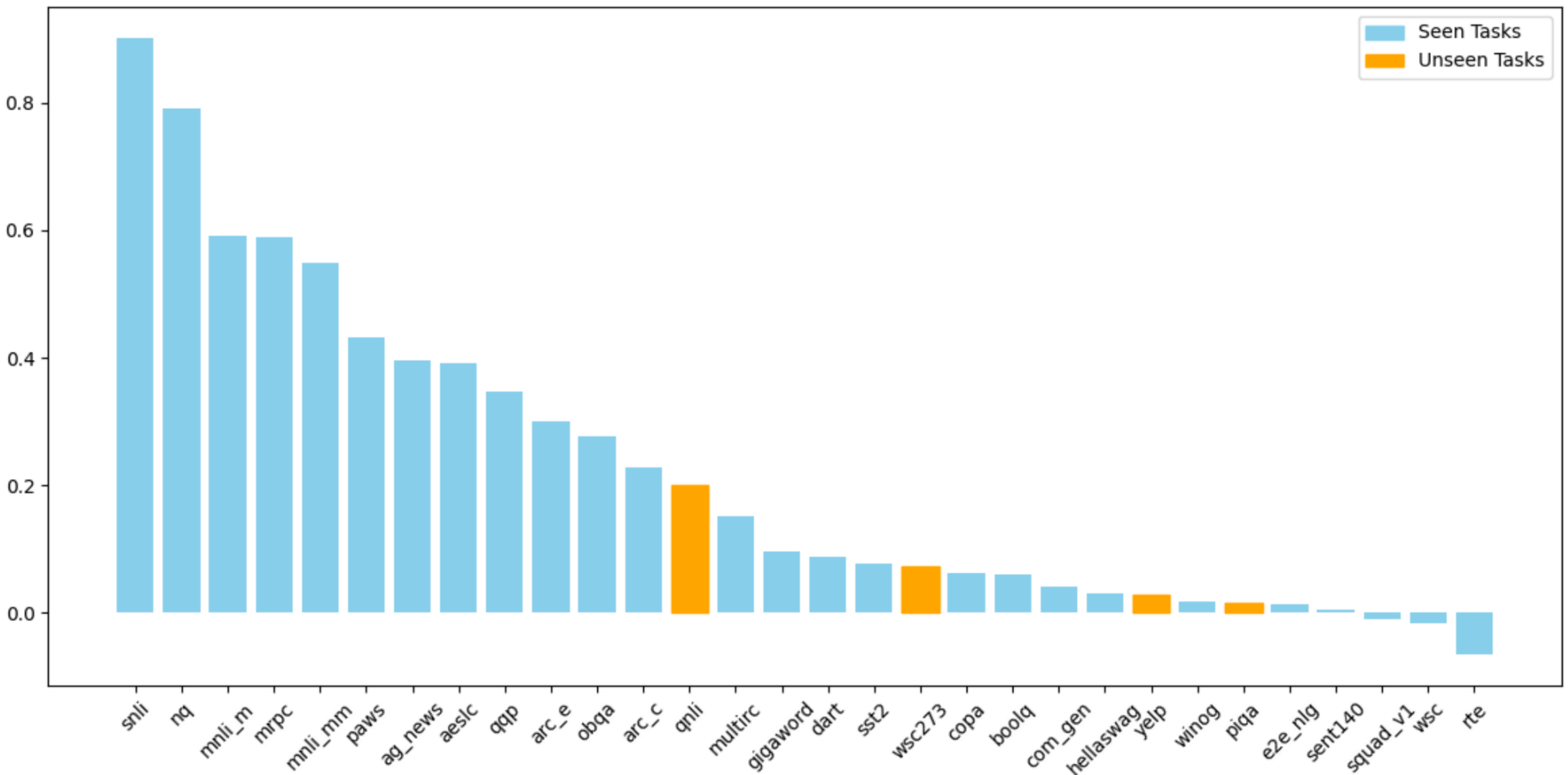}
  \vspace{-0.2cm}
  \caption{Performance gains of TDR over the random selection baseline.}
  \label{fig:performance_gains}
\end{figure*}
\subsection{Main Results}

Table \ref{main-results} presents the main results of our experiments. We report the average metrics for Close QA (CQA), Commonsense Reasoning (Comm.), Coreference (Coref.), NLI, Paraphrase (Para.), Reading Comprehension (RC), Sentiment (Sent.), Data-to-text (D2T), Summarize (Summ.). We adopt “Random” as a benchmark for comparison, which randomly selects examples for in-context learning evaluation. Dense retriever baselines include E5\cite{wang2022text}, SBERT\cite{reimers2019sentence}, EPR \cite{rubin2021learning} and LLM-R\cite{wang2023learning}.

\subsection{Universality and Performance Analysis of TDR}
Our method TDR is initially trained using feedback from LLaMA-7B. To validate its universality, we evaluate TDR on the aforementioned dataset in conjunction with larger language models GPT-Neo-2.7B\cite{black2021gpt} and LLaMA-13B without training. As shown in Table \ref{various-llms-results} the results reveal that our method TDR achieves average performance improvements of 0.5\% and 1.1\% over LLM-R, and surpasses the representative sparse retriever method BM25 by 7.9\% and 5.3\%, respectively. These findings underscore the versatility of our approach, which seamlessly integrates with diverse LLMs to enhance in-context learning capabilities by retrieving high-quality examples.

Notably, our method exhibits pronounced advantages in task types requiring semantically rich contexts, such as Paraphrase (Para.) and Reading Comprehension (RC) — where retrieved examples exhibit patterns closely aligned with the LLM’s response patterns. This performance gain is attributed to the higher-quality context retrieval enabled by our framework. Conversely, tasks of categories like Commonsense Reasoning (Comm.) and Data-to-text (D2T), where retrieved examples diverge significantly from the desired answer patterns and performance relies more heavily on the inherent reasoning capabilities of LLMs, the advantages of our method diminish. This phenomenon is corroborated by Table \ref{main-results} and Figure \ref{fig:performance_gains}. Table \ref{table:category_example} in the Appendix further illustrates this dichotomy by presenting representative retrieval examples from these two task types.

\subsection{Visualization of Training Effects}
\begin{figure}[t]
  \includegraphics[width=1\linewidth]{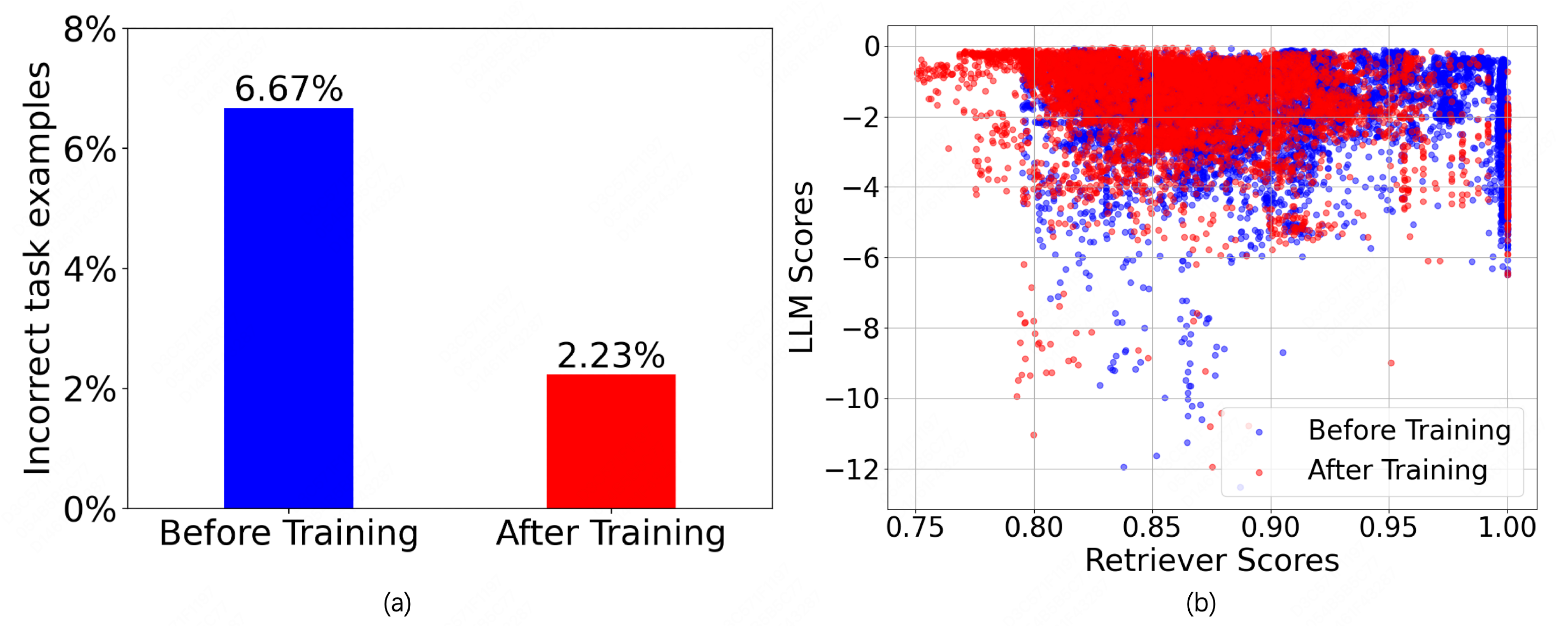}
  \caption {Visualization of Training Effects: (a) Proportion of Cross-Task Retrieval Before and After Training; (b) Correspondence Between Retrieved Examples and LLM Probabilities Before and After Training.}
    \label{fig:training_effects}
\end{figure}

To evaluate our correlation-enhanced loss, we analyze the retriever's performance before and after training using two metrics: (1) the proportion of retrieved examples from incorrect tasks, and (2) their impact on the language model's output probabilities. The results are shown in Figure~\ref{fig:training_effects}. In our setup, the retriever retrieves top-40 examples for 10,000 queries. The figure (a) shows the proportion of examples from incorrect tasks decreased from 6.67\% to 2.23\% after training, demonstrating our loss function's ability to focus on same-task examples. This aligns with our first objective.The figure (b) compares the output probabilities before (blue dots) and after (red dots) training. The red dots are more concentrated in the upper-left triangular region and overall higher, indicating that post-training examples lead to higher probabilities for the correct output $y$. This is expected, as retrieved examples should maximize $y$'s probabilities, aligning with our second objective.

\section{Conclusion}

In this work, we address two critical challenges in in-context learning (ICL) for large language models (LLMs): (1) difficulty in distinguishing cross-task data distributions and (2) underutilized fine-grained feedback from LLMs. To tackle these issues, we propose TDR, a novel framework that systematically enhances example retrieval for ICL through feedback-aware training and task-specific decoupling. The task-decoupled training strategy ensures precise retrieval of domain-relevant examples from multi-task datasets. Simultaneously, by designing a specialized correlation-enhanced loss function to model fine-grained LLM feedback, our method enables retrievers to learn patterns that retrieve better examples for LLMs.

Extensive experiments across 30 diverse NLP tasks demonstrate the superiority of TDR, achieving state-of-the-art performance over existing methods. Notably, our framework shows strong generalization capabilities, maintaining consistent gains on unseen tasks and across LLMs of varying~scales. These results validate that explicit modeling of LLM feedback and task-decoupled training strategy are crucial for unlocking~the~full~potential~of~ICL.


\section{Limitations}
The inherent feature discrepancies across different tasks presenting persistent challenges in developing task decoupling strategies. In our framework, TDR considers retrieval examples as two categories: belonging to the current task and not belonging to the current task, which may result in the ICL ability not benefiting from examples of similar tasks. More research remains necessary to develop adaptive penalty mechanisms that adjust penalty coefficients based on inter-task feature divergence magnitude, such as applying stronger regularization for tasks with significant feature disparities while reducing constraints for those with minimal discrepancies. 

Another limitation of our study is related to the utilization of high-quality examples retrieved during evaluation periods. Based on previous studies, we set the number of in-context examples to 8 and used it for a single round inference evaluation. However, the mutual coordination and influence among retrieval examples, as well as the way in which LLMs utilize these retrieval examples, such as using multiple rounds of evaluation instead, can be a promising direction for further exploration.

\bibliography{acl_latex}

\appendix

\section{A Comparative Analysis of KL-Divergence Loss Between This Method and Other Existing Approaches}
\textbf{REPLUG}:
    \begin{itemize}
        \item Directly minimizes $\mathcal{L}_{KL}(P_{LLM} \| P_{Retriever})$
        \item Faces instability due to distributional mismatch ($\sigma^2_{LLM}=2332$ vs $\sigma^2_{R}=2.58$)
    \end{itemize}
\textbf{LLM-R}
    \begin{itemize}
        \item Uses reward model to compress LLM probabilities ($\sigma^2_{Reward}=13.48$)
        \item Minimizes $\mathcal{L}_{KL}(P_{Reward} \| P_{Retriever})$ 
        \item Improves stability but introduces approximation error
    \end{itemize}
\textbf{Our TDR}:

Our approach selects candidates by directly leveraging LLM probabilities, establishing positive correlation without explicit distribution fitting.
    \begin{itemize}
        \item For low-probability $c_i$: Forces retriever to decrease $s(x,c_i)$
        \item For high-probability $c_j$: Increases $P_R(c_j|x)$ automatically due to $\sum_{n=1}^{m} P_R(c_n|x)=1$
    \end{itemize}
    
\section{Detailed experimental results}
\label{sec:appendix_detail_results}
Table \ref{table:detailed_results} shows detailed comparisons between our method and previous in-context example retrieval methods on 30 tasks. On 21 of the tasks, our method achieved state-of-the-art performance and 
achieved a 1.8\% improvement in average performance across all tasks, demonstrating the effectiveness and potential of this in-context example retriever paradigm.

\begin{table*}
\centering
\begin{tabular}{lccccccccc}
\hline
Dataset       & Zero-shot & Random & Kmeans & BM25 & $\text{E5}_\text{{base}}$ & SBERT & EPR  & LLM-R & Ours \\ \hline
AESLC         & 5.8       & 19.4   & 19.0   & 26.8 & 27.0   & 25.3  & 26.0 & 27.3  & 27.0 \\
AGNews        & 31.5      & 67.4   & 71.9   & 90.6 & 90.6   & 90.2  & 91.8 & 93.5  & 94.0 \\
ARC Challenge & 35.6      & 39.7   & 40.5   & 40.3 & 44.6   & 42.8  & 43.0 & 43.6  & 48.8 \\
ARC Easy      & 51.0      & 60.0   & 61.8   & 59.9 & 63.0   & 63.1  & 63.1 & 63.3  & 78.0 \\
BoolQ         & 64.7      & 70.0   & 69.0   & 74.7 & 72.4   & 73.9  & 74.8 & 75.1  & 74.2 \\
CommonGen     & 19.2      & 36.3   & 34.4   & 37.6 & 37.4   & 37.6  & 39.2 & 37.7  & 37.8 \\
COPA          & 66.0      & 80.0   & 85.0   & 78.0 & 83.0   & 82.0  & 82.0 & 84.0  & 85.0 \\
DART          & 22.9      & 52.0   & 46.6   & 55.9 & 54.7   & 54.4  & 56.2 & 57.2  & 56.6 \\
E2E NLG       & 34.6      & 52.7   & 46.4   & 54.5 & 51.8   & 50.2  & 53.6 & 54.7  & 53.4 \\
Gigaword      & 15.3      & 30.0   & 30.7   & 32.7 & 32.5   & 32.6  & 32.4 & 32.5  & 32.9 \\
HellaSwag     & 71.5      & 73.9   & 74.0   & 74.9 & 75.2   & 75.3  & 75.2 & 75.5  & 76.1 \\
MNLI (m)      & 35.8      & 46.3   & 44.2   & 50.1 & 44.5   & 50.8  & 59.9 & 70.2  & 73.7 \\
MNLI (mm)     & 35.6      & 48.1   & 45.4   & 48.3 & 44.7   & 49.3  & 61.5 & 72.0  & 74.5 \\
MRPC          & 69.1      & 49.5   & 38.0   & 61.8 & 41.2   & 52.7  & 55.9 & 75.3  & 78.7 \\
MultiRC       & 57.0      & 48.5   & 34.1   & 54.2 & 56.0   & 55.3  & 50.4 & 51.5  & 55.9 \\
NQ            & 0.3       & 21.5   & 22.6   & 37.6 & 39.3   & 39.4  & 39.2 & 39.1  & 38.5 \\
OpenBook QA   & 41.6      & 49.8   & 49.0   & 49.6 & 51.4   & 51.4  & 49.6 & 52.2  & 63.6 \\
PAWS          & 53.2      & 57.0   & 56.6   & 56.6 & 55.4   & 58.2  & 57.7 & 56.6  & 81.6 \\
PIQA          & 77.0      & 79.1   & 79.4   & 81.3 & 81.3   & 80.7  & 80.5 & 81.6  & 80.3 \\
QNLI          & 49.2      & 56.4   & 53.4   & 62.2 & 61.5   & 61.9  & 65.0 & 69.6  & 67.7 \\
QQP           & 57.7      & 63.4   & 63.3   & 79.8 & 77.5   & 81.3  & 81.7 & 82.6  & 85.4 \\
RTE           & 59.6      & 59.9   & 58.5   & 65.7 & 63.9   & 67.2  & 66.8 & 68.6  & 56.0 \\
Sentiment140  & 49.3      & 88.6   & 89.4   & 90.8 & 93.9   & 92.2  & 91.4 & 91.1  & 89.1 \\
SNLI          & 39.8      & 43.7   & 52.5   & 47.1 & 53.5   & 58.4  & 68.4 & 82.0  & 83.1 \\
SQuAD v1      & 2.1       & 64.1   & 62.3   & 61.2 & 60.8   & 61.6  & 64.3 & 57.3  & 63.5 \\
SST2          & 54.4      & 85.9   & 89.7   & 84.4 & 92.1   & 87.6  & 88.7 & 93.8  & 92.5 \\
Winogrande    & 62.0      & 66.7   & 66.5   & 67.5 & 66.9   & 66.5  & 66.5 & 68.1  & 68.0 \\
WSC           & 64.4      & 60.6   & 56.7   & 56.7 & 61.5   & 63.5  & 61.5 & 63.5  & 79.9 \\
WSC273        & 74.0      & 74.4   & 74.7   & 64.5 & 65.2   & 62.6  & 65.2 & 79.5  & 59.6 \\
Yelp          & 47.9      & 92.0   & 93.5   & 93.5 & 97.3   & 95.9  & 95.1 & 95.9  & 94.7 \\ \hline
Average       & 44.9      & 57.9   & 57.0   & 61.3 & 61.4   & 62.1  & 63.5 & 66.5  & 68.3 \\ \hline
\end{tabular}
\caption{\label{table:detailed_results}
Detailed experimental results for all 30 tasks of our main experiment.
}
\end{table*}

\section{Pattern analysis of retrieved examples from different task types}
As shown in Table \ref{table:category_example}, for the examples in the two lines above, which come from category Paraphrase (Para.) and Reading Comprehension (RC) respectively, retrieved examples exhibit patterns closely aligned with the patterns of queries and LLM's responses. For the examples in the two lines below, which come from category Commonsense Reasoning (Comm.) and Data-to-text (D2T) respectively, retrieved examples diverge significantly from the desired answer patterns and performance relies more heavily on the inherent reasoning capabilities of LLMs.

\section{Analysis of retrieval examples from other tasks}
\label{sec:appendix_other_task_analyse}
As shown in Table \ref{table:diff_task_case_analyse}, examples retrieved from other tasks have similar text content with the queries, but patterns and contents of the retrieved answers are significantly different from those required for the answer corresponding to the query, which makes distinguishing retrieval examples from different tasks an important factor limiting in-context learning performance of LLMs.

\begin{table*}
\centering
\begin{tabular}{ll}
\hline
Task name         & QQP                                                                                                                                                                                                                                                                                                       \\
Test Input        & \begin{tabular}[c]{@{}l@{}}"How will I contact a good hacker?" "How do l contact a hacker?" \\ Would you say that these questions are the same?\end{tabular}                                                                                                                                              \\
Test Answer       & \textbf{Yes}                                                                                                                                                                                                                                                                                              \\
Retrieved Example & \begin{tabular}[c]{@{}l@{}}"How will I contact a genuine hacker?" "How do l contact a   hacker?" \\ Would you say that these questions are the same? \textbf{Yes}\end{tabular}                                                                                                                                     \\ \hline
Task name         & BoolQ                                                                                                                                                                                                                                                                                                     \\
Test Input        & \begin{tabular}[c]{@{}l@{}}Tinker Bell (film series) -- A live-action film, with Reese Witherspoon playing \\ Tinker Bell and Victoria Strouse writing the script, is in the works.   \\ Can we conclude that are there going to be more tinkerbell movies?\end{tabular}                                  \\
Test Answer       & \textbf{Yes}                                                                                                                                                                                                                                                                                              \\
Retrieved Example & {\color[HTML]{3B3B3B} \begin{tabular}[c]{@{}l@{}}Tinker Bell (film series) -- A   live-action film, with Reese Witherspoon playing \\ Tinker Bell and   Victoria Strouse writing the script, is in the works. \\ Can we conclude that are there going to be any more tinkerbell movies? \textbf{Yes}\end{tabular}} \\ \hline
Task name         & COPA                                                                                                                                                                                                                                                                                                      \\
Test Input        & The horse bucked. What is the cause?                                                                                                                                                                                                                                                                      \\
Test Answer       & \textbf{The rider stroked the horse.}                                                                                                                                                                                                                                                                     \\
Retrieved Example & The rider fell to the   ground. What is the cause? \textbf{The bull bucked the rider.}                                                                                                                                                                                                                            \\ \hline
Task name         & DART                                                                                                                                                                                                                                                                                                      \\
Test Input        & \begin{tabular}[c]{@{}l@{}}Triple: Belgium, LANGUAGE, German language\\ What is a sentence that describes this triple?\end{tabular}                                                                                                                                                                       \\
Test Answer       & \textbf{German is the spoken language in Belgium.}                                                                                                                                                                                                                                                        \\
Retrieved Example & \begin{tabular}[c]{@{}l@{}}Triple: Belgium,  LANGUAGE, French language\\ What is a sentence that describes this triple? \\ \textbf{French is the spoken language in Belgium.}\end{tabular}                                                                                                                         \\ \hline
\end{tabular}
\caption{\label{table:category_example}
The bold texts are the ground-truth answers for the test inputs and retrieved candidates. These four examples belong to the category Paraphrase, Reading Comprehension, Commonsense Reasoning and Data-to-text respectively.
}
\end{table*}

\begin{table*}
\centering
\begin{tabular}{ll}
\hline
Task name         & DART                                                                                                                                                                                                                                                                                                                                                                                                                                                                                                                                                                                                                                                                                                                                                                                                                                                                                                                                                                                                                                                                                                                                                                                                                                                                                                                          \\
Test Input        & \begin{tabular}[c]{@{}l@{}}Triple: Clowns, priceRange, cheap; Clowns, familyFriendly, yes; Clowns, \\ near, Café Sicilia. What is a sentence that describes this triple?\end{tabular}                                                                                                                                                                                                                                                                                                                                                                                                                                                                                                                                                                                                                                                                                                                                                                                                                                                                                                                                                                                                                                                                                                                                         \\
Test Answer       & \textbf{A family friendly place is Clowns. It's cheap. It's near Café Sicilia.}                                                                                                                                                                                                                                                                                                                                                                                                                                                                                                                                                                                                                                                                                                                                                                                                                                                                                                                                                                                                                                                                                                                                                                                                                                               \\
Retrieved Context & \begin{tabular}[c]{@{}l@{}}Attributes: name = Clowns, priceRange = cheap, familyFriendly = yes, \\ near = Café Sicilia. Produce a detailed sentence about this restaurant.\end{tabular}                                                                                                                                                                                                                                                                                                                                                                                                                                                                                                                                                                                                                                                                                                                                                                                                                                                                                                                                                                                                                                                                                                                                       \\
Retrieved Answer  & \textbf{\begin{tabular}[c]{@{}l@{}}A newly-opened venue near Café Sicilia, Clowns offers cheap, family\\ -friendly dining.\end{tabular}}                                                                                                                                                                                                                                                                                                                                                                                                                                                                                                                                                                                                                                                                                                                                                                                                                                                                                                                                                                                                                                                                                                                                                                                      \\ \hline
Task name         & NQ                                                                                                                                                                                                                                                                                                                                                                                                                                                                                                                                                                                                                                                                                                                                                                                                                                                                                                                                                                                                                                                                                                                                                                                                                                                                                                                            \\
Test Input        & Question: who do you play as in halo 5? Answer:                                                                                                                                                                                                                                                                                                                                                                                                                                                                                                                                                                                                                                                                                                                                                                                                                                                                                                                                                                                                                                                                                                                                                                                                                                                                               \\
Test Answer       & \textbf{a Spartan}                                                                                                                                                                                                                                                                                                                                                                                                                                                                                                                                                                                                                                                                                                                                                                                                                                                                                                                                                                                                                                                                                                                                                                                                                                                                                                            \\
Retrieved Context & @5toSucceed @halo9 thank you. What is the sentiment of this tweet?                                                                                                                                                                                                                                                                                                                                                                                                                                                                                                                                                                                                                                                                                                                                                                                                                                                                                                                                                                                                                                                                                                                                                                                                                                                            \\
Retrieved Answer  & \textbf{Positive}                                                                                                                                                                                                                                                                                                                                                                                                                                                                                                                                                                                                                                                                                                                                                                                                                                                                                                                                                                                                                                                                                                                                                                                                                                                                                                             \\ \hline
Task name         & MultiRC                                                                                                                                                                                                                                                                                                                                                                                                                                                                                                                                                                                                                                                                                                                                                                                                                                                                                                                                                                                                                                                                                                                                                                                                                                                                                                                       \\
Test Input        & \begin{tabular}[c]{@{}l@{}}\{ \{ lang \} \} centers on a man who roams the street night after night. \\ Hidden under his hat and rain jacket he strives for one goal : \\ to find the culprit - the one whom he can make responsible for his suffering . \\ If he wanted to , he could confront him , but he lacks the audacity to do so . \\ He considers suicide , but his courage fails him once again . \\ The options do not appear to present him with a way out and would not \\ personally satisfy him . Finley blames not himself , but only others . \\ In this case he looks to his girlfriend , Violet . He drowns Violet in the bath \\ whilst giving her a massage , Which had become a common ritual for them . \\ On one hand he does this out of malice , on the other to be close to her just one \\ more time. Through this action he wishes to break the growing distance he has \\ come to feel between them , though the actual outcome is the infliction of the \\ greatest possible loneliness , as he turns into a monster . Finley only realizes \\ with hindsight that his misdeeds far surpass those of Violet . \\ Question: "What was Finley doing with Violet before he killed her?" \\ Response: "They were in bed together" \\ Does the response correctly answer the question?\end{tabular} \\
Test Answer       & \textbf{No}                                                                                                                                                                                                                                                                                                                                                                                                                                                                                                                                                                                                                                                                                                                                                                                                                                                                                                                                                                                                                                                                                                                                                                                                                                                                                                                   \\
Retrieved Context & \begin{tabular}[c]{@{}l@{}}Write a short summary for this text: or how about a girl who is equally obsessed \\ with this guy even though he continually tells her he 's dangerous , could \\ inadvertently kill her and treats her as if she were a child ? this same girl becomes \\ so depressed when her boyfriends breaks up with her that she begins to take \\ risks, some seemingly suicidal , because such behavior summons visions of him.\end{tabular}                                                                                                                                                                                                                                                                                                                                                                                                                                                                                                                                                                                                                                                                                                                                                                                                                                                              \\
Retrieved Answer  & \textbf{some scholars find disturbing elements in twilight books}                                                                                                                                                                                                                                                                                                                                                                                                                                                                                                                                                                                                                                                                                                                                                                                                                                                                                                                                                                                                                                                                                                                                                                                                                                                             \\ \hline
\end{tabular}
\caption{\label{table:diff_task_case_analyse}
Retrieved examples from other tasks. The bold texts are the ground-truth answers for the test inputs and retrieved candidates. 
}
\end{table*}

\end{document}